%% file: lagrangian-ot.tex
\documentclass[accepted]{uai2024} %
\usepackage[american]{babel}

\input{paper_preamble}
\usepackage{algorithm}
\usepackage{algorithmic}

\usepackage{natbib} %
\usepackage{mathtools} %
\usepackage{booktabs} %
\usepackage{tikz} %

\title{Neural Optimal Transport with Lagrangian Costs}

\author[1,3]{Aram-Alexandre~Pooladian}
\author[2,3]{Carles~Domingo-Enrich}
\author[3]{Ricky~Tian~Qi~Chen}
\author[3]{Brandon~Amos}
\affil[1]{%
    Center for Data Science\\
    New York University}
\affil[2]{%
    Courant Institute of Mathematical Sciences\\
    New York University}
\affil[3]{FAIR, Meta}
  
\begin{document}
\maketitle

\begin{abstract}
  We investigate the optimal transport problem between probability measures
  when the underlying cost
  function is understood to satisfy a \emph{least action principle},
  also known as a \emph{Lagrangian} cost.
  These generalizations are useful when connecting observations
  from a physical system where the transport dynamics are influenced
  by the geometry of the system, such as obstacles (\eg, incorporating barrier functions in the Lagrangian), and allows practitioners to incorporate \emph{a priori} knowledge of the
  underlying system such as non-Euclidean geometries (\eg, paths must be circular).
  Our contributions are of computational interest, where
  we demonstrate the ability to efficiently compute geodesics and
  amortize spline-based paths, which has not been done before, even in low dimensional problems. Unlike prior work, we also output the resulting \emph{Lagrangian optimal transport map} without requiring an ODE solver.
  We demonstrate the effectiveness of our formulation on low-dimensional examples taken from  prior work. 
  The source code to reproduce our experiments is available at
  \url{https://github.com/facebookresearch/lagrangian-ot}.
\end{abstract}
\begin{figure}[t]
  \begin{tikzpicture}[every node/.style={inner sep=0,outer sep=0}]
    \node[anchor=south west] at (0, 0) {\includegraphics[width=0.49\linewidth]{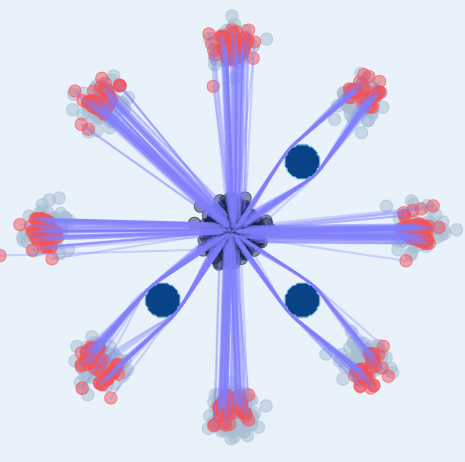}};
    \node[anchor=south west,align=left] at (0.0, 0) {\small (a) Obstacles};
  \end{tikzpicture}%
  \begin{tikzpicture}[every node/.style={inner sep=0,outer sep=0}]
    \node[anchor=south west] at (0, 0) {\includegraphics[width=0.49\linewidth]{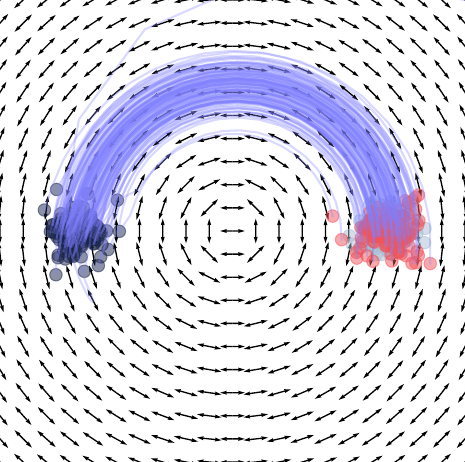}};
    \node[anchor=south west,fill=white,align=left,fill opacity=0.9] at (0.0, -.5mm) {%
      \small
      (b) Circular Geometry\vspace{-.5mm}};
  \end{tikzpicture} \\
  \noindent {\small samples (\cblock{26}{37}{75} source
  \cblock{167}{190}{211} target \cblock{242}{84}{92} push-forward)
  \hfill
  \cblock{168}{168}{254} transport paths}
  \caption{Optimal transport paths with Lagrangian costs
    on the obstacles setting from \citet{liu2022deep}
    and circular geometry from \citet{scarvelis2022riemannian}.}
  \label{fig:synthetic-examples_intro}
\end{figure}

\section{Introduction}
Computational efforts in optimal transport traditionally revolve around the squared-Euclidean cost $\tfrac12\|x-y\|^2$. This cost has a connection to convex functions via Brenier's theorem \citep{brenier1991polar}, and has allowed for both
numerical analysts \citep{jacobs2020fast} and machine learning
researchers
\citep{bunne2022proximal,amos2022amortizing,korotin2019wasserstein} to
push the boundaries of computational optimal transport in recent
years.  This connection has also been influential in domains such as
economics and statistics
\citep{carlier2016vector,chernozhukov2017monge}, high-energy particle
physics \citep{manole2022background}, computational biology
\citep{schiebinger2019optimal,bunne2021learning,bunne2022supervised},
computer vision \citep{feydy2017optimal}, among others.

However, there is little reason
practitioners should \emph{default} to this cost in their
applications, where often they know that paths will not be straight lines, or have obstacles that must be avoided.  The purpose of this paper is to provide a computational
framework that allows practitioners to enforce transport with
more general costs that can incorporate such geometries.  To this end, our goal is to
numerically solve the optimal transport problem when the underlying
cost of displacement is governed by a \emph{least action principle}.
For two points $x,y \in \R^d$, the \emph{displacement cost} $c(x,y)$ is
\begin{align}\label{eq:cost_intro}
  c(x,y) = \inf_{\gamma\in\cC(x,y)} \left\{ \int_0^1 \cL(\gamma_t, \dot{\gamma}_t) \dd t \right\},
\end{align}
where $\cC(x,y)$ is the set of smooth, time dependent curves
$\gamma$ that connect $x$ and $y$ such that $\gamma_0=x$ and $\gamma_1=y$,
and
  $\cL: \R^d\times \R^d \rightarrow \R\cup\{+\infty\}$
is the \emph{Lagrangian function} which ultimately governs the
cost of transport.
The Lagrangian takes as arguments the position of the curve $\gamma_t\in\R^d$
at time $t$ and the velocity at that point $\dot\gamma_t\defeq \tfrac{\dd\gamma_t}{\dd t}$.
This definition is inspired by Lagrangian mechanics: equations of motion that are based on
energies of a system, rather than forces. As outlined in
\citet[Chapter 7]{Vil08}, and briefly discussed in
\cref{sec:background_lagot}, this notion of cost can be lifted to the space of
probability measures, instead of just being between two fixed points
over a space.

If $\cL(x,v) = \tfrac12\|v\|^2$, then $c(x,y)$ recovers the
squared-Euclidean distance of transport
(\cf~\citet{benamou2000computational}).
However, the Lagrangians that
we consider are more general. They not only impact the transport destination, but also the optimal path (instead of just straight lines). Examples include 
\begin{description}
    \item 1) \emph{potential energy terms} (see \cref{ex:kinetic_potential}), with
 $$\cL(x,v) = \tfrac12\|v\|^2 - U(x)\,,$$
\item 2) \emph{position-dependent costs} (see \cref{ex:geodesic}), with
$$\cL(x,v) = \!\tfrac12\|v\|^2_{A(x)} \defeq \tfrac12 v^\top A(x) v\,,$$
\end{description}
where $U : \R^d \to \R$ is a potential function, and $A : \R^d \to \mathbb{S}^{d}_{++}$ is a (positive definite) matrix-valued function.

\Cref{fig:synthetic-examples_intro} contains two Lagrangian optimal transport problems. In \cref{fig:synthetic-examples_intro}(a), smooth potential functions act as obstacles between the source Gaussian and the
8-Gaussian mixture. In \cref{fig:synthetic-examples_intro}(b), the cost of
displacement is lowest along circular trajectories. We stress that,
despite being non-standard notions of cost, there still exists
an optimal transport map (see \cref{eq:transport_map_c}) expressed
explicitly as the minimizer to an optimization problem, and the paths are unique. These notions of cost have appeared in a variety of works (see e.g., \cite{koshizuka2022neural,scarvelis2022riemannian,liu2022deep}), though none of these approaches provide \emph{deterministic} mappings (i.e., source-to-target maps in one function evaluation), nor do they provide optimal paths.

\textbf{Main contributions.}
We aim to fill this gap in the literature on computational optimal transport, where
the cost function follows \cref{eq:cost_intro} with potential energies or position-dependent costs, and the measures
are in an underlying continuous space. Our two main goals, which go hand-in-hand, are to:
\begin{description}
  \setlength\itemsep{0.1em}
    \item (1) Compute the Lagrangian optimal transport maps,
    \item (2) Compute the resulting \emph{paths} for these maps.
\end{description}
We want to emphasize that, to the best of our knowledge, these approaches have not been considered in the machine learning literature related to  \emph{unregularized} optimal transport. Indeed, since the cost function \cref{eq:cost_intro} is itself a minimization problem, the resulting OT problem (see \cref{sec:background}) is a \emph{bi-level} optimization problem, wherein lies the difficulty of these general costs. Our work acts as a first step to tackling these optimization problems in a principled manner.

Borrowing inspiration from the existing literature, we consider two categories of optimal transport problems: i) transport between a \emph{pair} of probability measures
$(\mu,\nu)$, and ii) transport between \emph{consecutive pairs} of
measures $\{(\rho_i,\rho_{i+1}\}_{i=0}^{K-1}$.
In i), we assume the practitioner is interested in modeling physical
systems, and has access to their Lagrangian of interest, either
through the potential function $U$ or position-dependent metric $A$,
and wants to know the optimal displacement and cost between $\mu$ and
$\nu$.
For ii), the practitioner has access to samples from $K$ probability
measures, which they believe to be traversing optimally under some
underlying Riemannian metric which is not known.
We \emph{learn} this Riemannian metric to uncover the geometry
of the space.
We stress that both modifications allow for the
practitioner to employ in-domain knowledge to the cost function, as
opposed to the squared-Euclidean cost, which remains
information agnostic.

Our approach involves parameterizing the Lagrangian optimal transport maps and paths using neural networks.
The non-standard cost leads to two computational challenges for obtaining
1) the displacement cost \cref{eq:cost_intro} and minimizing path,
2) the $c$-transform of the Lagrangian cost.
We overcome both of these by using amortized optimization
(\eg as in \citet{amos2023tutorial}) to obtain approximate solutions.
In the two tasks we consider (Lagrangian optimal transport between two measures, and Riemannian metric learning through a sequence of pairs of measures), we outperform existing baselines on data taken from the respective papers.

\section{Background on optimal transport }\label{sec:background}
\subsection{Kantorovich primal-dual problems and optimal transport mappings}\label{sec:background_kantdual}
Optimal transport can be written as several equivalent infinite-dimensional optimization problems, which we outline below under mild conditions. We refer the interested reader to \citet{San15} or \citet{Vil08} for a more detailed discussion.
Let $\mu \in \cP(\cX)$ and $\nu \in \cP(\cY)$ be two probability
measures defined on $\cX$ and $\cY$, respectively, which are complete,
separable metric spaces (for simplicity, one can consider $\R^d$
endowed with the Euclidean metric). Let $c : \cX \times \cY \to \R$ be
a lower semicontinuous, real-valued cost function (for simplicity, one
can consider any bounded convex cost function).

The \emph{primal
(Kantorovich) formulation}, attributed to \citet{Kan42}, is given by
\begin{align}\label{eq:kantprimal_generalc}
    \text{OT}_c(\mu,\nu) \defeq \inf_{\pi \in \Gamma(\mu,\nu)} \iint_{\cX \times \cY} c(x,y) \dd \pi(x,y)\,,
\end{align}
where $\Gamma(\mu,\nu) \subset \cP(\cX \times \cY)$ is the set of
transportation couplings between $\mu$ and $\nu$ \ie, $\pi \in
\Gamma(\mu,\nu)$ if
\begin{align}
    \int_\cY \dd \pi(x,y) = \dd \mu(x)\,, \quad  \int_\cX \dd \pi(x,y) = \dd \nu(x)\,.
\end{align}

Under our specifications on the cost function, an equivalent
optimization problem called the \emph{dual (Kantorovich) formulation},
\cf \citet[Theorem 5.10]{Vil08}, is
\begin{align}\label{eq:kantdual_generalc}
  \hspace*{-1mm}
  \text{OT}_c(\mu,\nu) =
  \hspace*{-1mm}
  \sup_{g \in L^1(\nu)} \int g^c(x) \dd \mu(x) + \int g(y) \dd \nu(y)\,,
  \hspace*{-1mm}
\end{align}
where $L^1(\nu)$ is the set of integrable functions with respect to $\nu$, and $g^c$ is the \emph{$c$-transform of $g$}, written
\begin{align}
  \label{eq:ctransform}
  \hspace*{-3mm}
  g^c(x) \defeq \inf_{y \in \cY} J(y; x)\ \text{where}\ J(y;x)\defeq c(x,y) - g(y)\,.
  \hspace*{-1mm}
\end{align}
When attained, the minimizer of \cref{eq:ctransform} is
\begin{align}\label{eq:transport_map_c}
   \hat{y}(x;c,g) \defeq \argmin_{y \in \cY}\{c(x,y) - g(y)\}\,.
\end{align}
When the supremum in \cref{eq:kantdual_generalc} is attained,
we write $\hat{g}$ as the maximizer, called the optimal Kantorovich
potential. We define the \emph{optimal transport map} associated to the cost $c$ as
the minimizer $\hat{y}(\cdot;c, \hat{g})$, which is
\cref{eq:transport_map_c} applied to the optimal Kantorovich
potential. Given $x \in \cX$, $\hat{y}(x;c,\hat{g})$ corresponds to
the optimal displacement from $\mu$ to $\nu$.

\subsection{Lagrangian Optimal Transport (LOT)}\label{sec:background_lagot}
We now suppose our probability measures exist on compact subsets $\cX = \cY \subseteq \R^d$.
We associate the cost of displacing $x$ to $y$ with an \emph{action}
that is to be minimized over a time horizon $[0,1]$. Borrowing
terminology from physics, these actions will take the form of
\emph{Lagrangian} functionals, which are functions that depend on the
position of a curve $\gamma_t$, its velocity, $\dot\gamma_t$, and time
$t \in [0,1]$;
\begin{align}
  \label{lagrangian-def-gamma}
    (\gamma_t,\dot\gamma_t) \mapsto \cL(\gamma_t,\dot\gamma_t)\,,
\end{align}
where curves in $\cC$ are understood to be smooth
and absolutely continuous curves over $\R^d$,
indexed by time in $[0,1]$, \cf \citet[Chapter 7]{Vil08}.
The Lagrangian induces an \emph{action} or \emph{energy} $E$ on curves defined
by
\begin{align}
  \label{eq:lag_energy}
  E(\gamma; x, y) = \left\{ \int_0^1 \cL(\gamma_t,\dot\gamma_t)\dd t \right\}\,.
\end{align}
The \emph{cost of displacement} is
then given by
\begin{align}
  \label{eq:lag_cost}
  c(x,y) = \!\!\!\inf_{\gamma \in \cC(x,y)} E(\gamma; x, y)\,. %
\end{align}
Though initially defined between two points on the manifold, this cost
can be appropriated ``lifted'' to the space of probability measures,
resulting in what is known as \emph{Lagrangian Optimal Transport} (LOT). 
Indeed, under mild assumptions on $\cL$, the generalized notion of
transport vis-{\`a}-vis minimizers to \cref{eq:ctransform} is
defined. A thorough discussion is found in \citet[Chapter 7]{Vil08},
specifically Theorem 7.21 and Remark 7.25.
The following conditions are sufficient for \cref{eq:lag_cost} to
define a valid notion of transport: $\cL$ is twice continuously
differentiable and strictly convex in $v$, with $\nabla^2_v \cL \succ 0$
everywhere, and $\cL$ does not depend (explicitly) on $t$.
These conditions are satisfied in \emph{all} our problem considerations. Thus, when $c$ is a cost of the form \cref{eq:lag_cost}, we refer to the Lagrangian optimal transport map (or LOT map) as the minimizer to \cref{eq:transport_map_c} under this cost.

\begin{remark}
  For simplicity, we present the background for manifolds
  $(\R^d,g)$ where $g$ is potentially a non-Euclidean metric.
  These same discussions hold when we instead consider a
  general smooth Riemannian manifold $\cM$ and its associated
  metric $g$; \cf \eg, \citet{feldman2002monge}. %
\end{remark}

\begin{example}[Euclidean distances, \cf \citet{benamou2000computational}]
  \label{ex:euclidean_kinetic}
  The squared Euclidean distance is recovered, \ie, $c(x,y)=\|x-y\|_2^2$,
  by taking the Lagrangian as the kinetic energy:
  \begin{align}\label{eq:kinetic_eq}
    \cL(\gamma_t,\dot\gamma_t,t) = \tfrac12 \|\dot\gamma_t\|^2\,.
  \end{align}
  Indeed, $\cL$ is twice differentiable with $\nabla^2_v \cL = I \succ 0$, which satisfies our conditions.
\end{example}
\begin{example}[Obstacles and other potential functions]
  \label{ex:kinetic_potential}
One can add a potential function (not to be confused with \emph{Kantorovich} potentials from \cref{sec:background_kantdual})
$U : \R^d \to \R$, to the kinetic energy, resulting in the Lagrangian
\begin{align}\label{eq:lagrangian_eq}
    \cL(\gamma_t, \dot{\gamma}_t) = \frac{1}{2}\|\dot{\gamma}_t\|^2  - U(\gamma_t)\,.
\end{align}
Again, $\nabla^2_v \cL = I \succ 0$. We require $U$ to be sufficiently smooth in order for $\cL$ to be twice continuously differentiable.
The function $U$ provides a way of specifying how ``easy''  or ``hard'' it is
to pass through regions of the space.
This includes the obstacles as in
\cref{fig:synthetic-examples_intro}(a) and \cref{fig:synthetic-examples}
where the potential takes low values and prevents the
paths from crossing them.
\end{example}

\begin{algorithm*}[t]
\caption{Neural Lagrangian Optimal Transport (NLOT)}
  \begin{algorithmic}
    \STATE \textbf{inputs:} measures $\mu$ and $\nu$, Kantorovich potential $g_\theta$,
    $c$-transform predictor $y_\zeta$, and spline predictor $\varphi_\eta$
    \WHILE {unconverged}
    \STATE sample batches $\{x_i\}_{i=1}^N\sim\mu$ and $\{y_i\}_{i=1}^N\sim\nu$
    \STATE obtain the amortized $c$-transform predictor $y_\zeta(x_i)$ for $i \in [N]$
    \STATE fine-tune the $c$-transform by numerically solving \cref{eq:transport_map_c},
    warm-starting with $y_\zeta(x_i)$
    \STATE update the potential with gradient estimate of $\nabla_\theta \ell_{\text{dual}}$
    (\cref{eq:kantdual_neural_derivative})
    \STATE update the $c$-transform predictor $y_\zeta$ using a gradient
        estimate of \cref{eq:ctransform-amor}
    \STATE update the spline predictor $\varphi_\eta$ using a gradient
        estimate of \cref{eq:spline-amor}
    \ENDWHILE
    \STATE \textbf{return} optimal parameters $\theta$, $\phi$, $\eta$
  \end{algorithmic}
  \label{alg:lagrangian-ot}
\end{algorithm*}

\begin{example}[Squared geodesic distances on Riemannian manifolds]
  \label{ex:geodesic}
  \Cref{ex:euclidean_kinetic} can be extended to non-Euclidean
  manifolds.
  In $\R^d$, the metric at a point $x \in \R^d$ is given by the
 inner product $\langle u , v \rangle_x \defeq \langle u, A(x) v
  \rangle$ for any $u,v \in \R^d$, for $A(\cdot) : \R^d \to
  \mathbb{S}^{d}_{++}$ positive-definite,
  giving the Lagrangian
  \begin{align}\label{eq:lagrangian_eq_ametric}
    \cL(\gamma_t,\dot{\gamma}_t; A) = \frac{1}{2}\|\dot\gamma_t\|^2_{A(\gamma_t)}\,.
  \end{align}
  Here, $\nabla_v^2 \cL = A(\gamma_t)$, so we require $A(\cdot) \succ 0$ to satisfy the criteria of Theorem 7.21 and Remark 7.25 from \citet{Vil08}.
\end{example}

\Cref{ex:geodesic} shows the circular
geometry in \cref{fig:synthetic-examples}(a)
where the metric is
given by the positive-definite matrix
\begin{align}\label{eq:circle_path}
    A(x) = \begin{pmatrix}
        \frac{x_1^2}{\|x\|^2} & 1 - \frac{x_1x_2}{\|x\|^2} \\
        1 - \frac{x_1x_2}{\|x\|^2} & \frac{x_2^2}{\|x\|^2}
    \end{pmatrix} \,.
\end{align}

\section{Lagrangian OT between two measures via neural networks}
\label{sec:neural_lagot}

We first focus on computationally solving for the Kantorovich dual
in \cref{eq:kantdual_generalc} between two measures $\mu \in \cP(\cX)$
and $\nu \in \cP(\cY)$ when the cost function is of the
form \cref{eq:lagrangian_eq} or \cref{eq:lagrangian_eq_ametric}.
All components of the Lagrangian are known, \ie, the Lagrangian potential $U$
or the underlying metric $A(\cdot)$ is known, and
we assume access to samples from $\mu$ and $\nu$.
The Kantorovich potential $g\in L^1(\nu)$ in \cref{eq:kantdual_generalc}
is a function $g: \cY\rightarrow \R$.
We present a detailed explanation below; \Cref{alg:lagrangian-ot} summarizes our solution.

We follow recent neural optimal transport methods,
\eg, \citet{taghvaei20192,makkuva2020optimal,korotin2019wasserstein,fan2021scalable,amos2022amortizing},
and represent the Kantorovich potential as a neural network $g_\theta$ with parameters $\theta$.
With this parameterization, we recast \cref{eq:kantdual_generalc} as
$\max_\theta \ell_{\rm dual}(\theta)$ where
\begin{align}
  \label{eq:kantdual_neural}
  \ell_{\rm dual}(\theta)\defeq\int g_{\theta}^{c}(x) \dd \mu(x) + \int g_{\theta}(y) \dd \nu(y)\,
\end{align}
and the $c$-transform $g_\theta^c$ incorporates the
Lagrangian function.
We optimize \cref{eq:kantdual_neural} via gradient descent, where the derivative in the parameters is 
\begin{equation}
  \label{eq:kantdual_neural_derivative}
  \nabla_\theta \ell_{\rm dual}(\theta) = \int \nabla_\theta g_{\theta}^{c}(x) \dd \mu(x) + \int \nabla_\theta g_{\theta}(y) \dd \nu(y),
  \small
\end{equation}
and $g_\theta^c$ is differentiated with
\emph{Danskin's envelope theorem} \citep{danskin1966theory,bertsekas1971control}, \ie,
\begin{equation}
  \label{eq:g_derivative}
  \nabla_\theta g_\theta^c(x) = \nabla_\theta J(\hat y(x); x, c, g_\theta)
      = -\nabla_\theta g_\theta(\hat y(x))\,.
\end{equation}

We follow \citet{taghvaei20192}, as well as other OT work based on neural networks, and
approximate \cref{eq:kantdual_neural} and \cref{eq:kantdual_neural_derivative}
with Monte-Carlo estimates of the integrals as they are
not computable in closed-form.
Computing these estimates still requires overcoming the following
challenges:

\begin{challenge}[Computing the $c$-transform]
\label{challenge:ctransform}
Estimating \cref{eq:kantdual_neural} and \cref{eq:kantdual_neural_derivative}
require obtaining the $c$-transform $g_\theta^c$ and the corresponding
minimizing point $\hat y(x; c, g)$. This requires solving the
optimization problem in \cref{eq:ctransform} for \textbf{every} $x$,
which does not have a closed-form solution.
\end{challenge}

Prior OT work for the squared-Euclidean cost settings had to overcome a similar challenge when the
$c$-transform becomes the Fenchel or convex conjugate operation:
\citet{taghvaei20192,korotin2021neural} use numerical solvers such as L-BFGS, Adam,
and other gradient-based methods,
\citet{makkuva2020optimal,korotin2019wasserstein,korotin2021neural} use an amortized approximation,
and \citet{amos2022amortizing} combines the amortized approximation with
a numerical solver.
For $c$-transforms, \citet{fan2021scalable} uses an amortized approximation
to overcome \cref{challenge:ctransform}.

We follow these works and overcome \cref{challenge:ctransform}
by amortizing the solution to \cref{eq:ctransform}.
This involves parameterizing an approximate $c$-transform
map $\hat y_\zeta\approx \hat y$ that we learn
with a regression-based loss
\begin{equation}
  \label{eq:ctransform-amor}
  \min_\phi \int \|\hat y(x)-y_\zeta(x)\| \dd \mu(x)\,.
\end{equation}

The conjugation model $\hat y_\zeta$ is only an approximation
and may be inaccurate as the potential $g_\theta$ changes
during training. An inaccurate approximation to the $c$-transform
results in a poor approximation to the objective in
\cref{eq:kantdual_neural}; to improve it, we follow
\citet{amos2022amortizing} and fine-tune the $c$-transform
prediction with a few steps of L-BFGS to solve \cref{eq:transport_map_c}, and warm-start it with the amortized prediction.

\begin{challenge}[Computing the cost $c$]
  \label{challenge:cost}
  Evaluating the Lagrangian cost $c$ that arises in the $c$-transform
  in \cref{eq:ctransform,eq:ctransform-amor} involves solving
  the optimization problem in \cref{eq:lag_cost} over paths.
  While closed-form solutions exist for simple manifolds,
  \eg, straight paths on Euclidean space or great arcs on spherical
  manifolds, the more general settings we consider
  do not admit closed-form solutions and need to be numerically solved.
\end{challenge}

Computationally representing paths and solving for Riemannian geodesics and
Lagrangian paths in \cref{eq:lag_cost} outside of the context
of optimal transport is an active research area.
We follow \citet{beik2021learning,stochman} and parameterize the space
of paths between $x$ and $y$ with a cubic spline
$\gamma_\varphi(x,y)$ (where the parameters are $\varphi$).
This spline parameterization transforms the optimization problem in
\cref{eq:lag_cost} to an optimization problem over the continuous-valued
parameters of the spline as
\begin{align}
  \label{eq:lag_min_spline}
  \varphi^\star(x,y)\defeq\argmin_{\varphi\in\Phi(x,y)} E(\varphi; x, y)
\end{align}
where
  $E(\varphi; x, y)\defeq\left\{ \int_0^1 \cL((\gamma_\varphi)_t,(\dot\gamma_\varphi)_t)\dd t \right\}$
and $\Phi(x,y)$ is the space of cubic splines between $x$ and $y$; see \cref{app:paths} for more details.

Solving \cref{eq:lag_min_spline} for every $c$-transform within
every evaluation for the OT cost is computationally intractable,
so we propose to amortize the path computation using
objective-based amortization as in \citet{amos2023tutorial}.
Given points $x$ and $y$, we parameterize the spline amortization
model with $\varphi_\eta(x,y)\approx\varphi^\star(x,y)$, where $\eta$ represents the weights of a neural network. We train $\varphi_\eta$ to compute the paths necessary for the Lagrangian cost, \ie,
\begin{equation}
  \label{eq:spline-amor}
  \min_\eta \int E(\varphi_\eta; x, \hat y(x))\dd \mu(x)\;.
\end{equation}

\subsection{Experiments}
\begin{figure*}[t]
  \centering
  \begin{tikzpicture}[every node/.style={align=left,anchor=south west, inner sep=0}]
    \node at (0,0) {\includegraphics[width=0.9\textwidth]{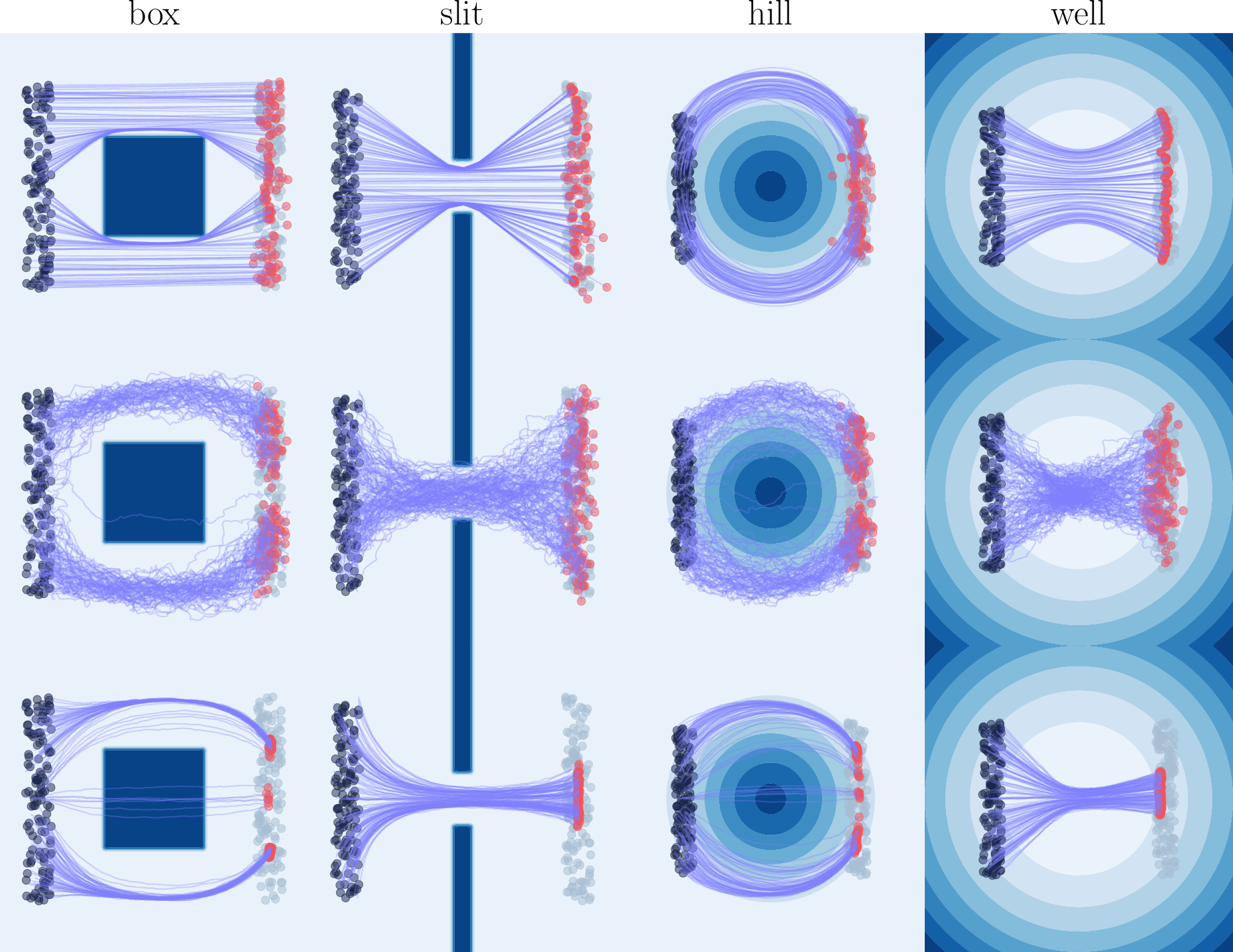}};
    \node at (0,11) {NLOT (\cref{alg:lagrangian-ot})};
    \node at (0,7.4) {NLSB (stochastic paths)};
    \node at (0,3.5) {NLSB (expected path)};
    \node[fill=white,fill opacity=0.8] at (0,0) {%
      contours indicate Lagrangian potential $U(x)$
      {\color{gray}(darker color=higher value)}};
  \end{tikzpicture} \\[2mm]
  \cblock{26}{37}{75} source samples \hspace{2mm}
  \cblock{167}{190}{211} target samples \hspace{2mm}
  \cblock{242}{84}{92} push-forward samples \hspace{2mm}
  \cblock{168}{168}{254} transport paths
  \caption{Paths on the Neural Lagrangian Schr{\"o}dinger Bridge (NLSB) datasets \citep{koshizuka2022neural}.}
  \label{fig:synthetic-examples}
\end{figure*}
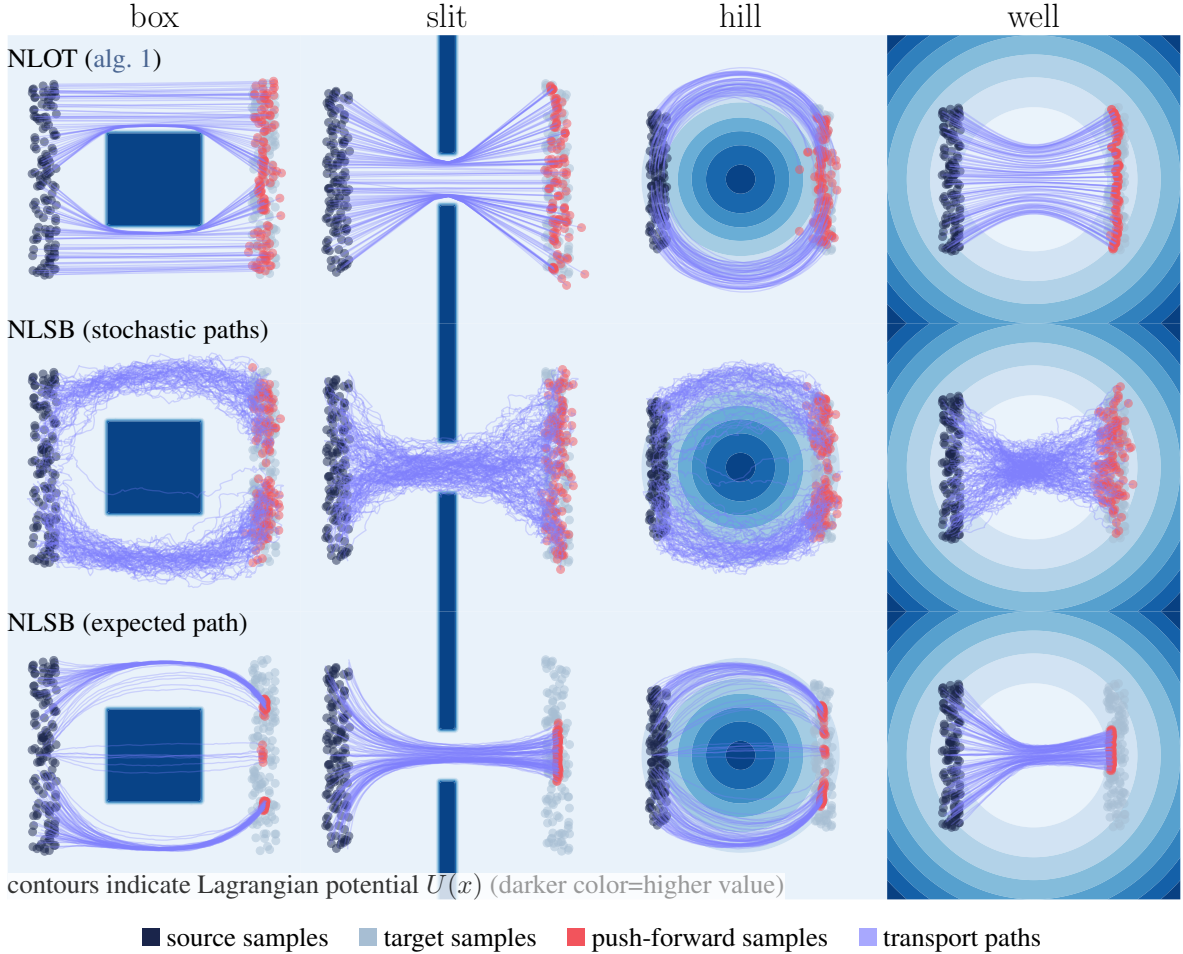

We consider Lagrangians of the form
\begin{align}
    L(x,v) = \tfrac12\|v\|^2 - U(x)\,,
\end{align}
with $U : \R^d \to \R$, where $U$ can take the form of a barrier that distorts the transport path from being a straight line. For modeling hard constraints (recall the three circular obstacles in \cref{fig:synthetic-examples_intro}(a)), we model $U$ as a smooth (but sharp) barrier function, which preserves smoothness in $L$. Precise definitions of the functions are deferred to \cref{app:potential_data}. In \cref{fig:synthetic-examples}, we learn
the optimal transport map between two measures with a box or slit constraint, or a hill or a well. 
In \cref{fig:synthetic-examples_intro}(a), we learn the optimal transport map between a Gaussian and a Gaussian mixture, with barriers (three circular barriers).
A single training run for all of the experimental settings takes
approximately 1-3 hours on our NVIDIA Tesla V100 GPU.

Our examples in \cref{fig:synthetic-examples} are taken from \cite{koshizuka2022neural}, and is our main benchmark for this task. Their approach is based on learning Schrodinger Bridges with neural networks, called NLSB (Neural Lagrangian Schrodinger Bridge). Therein, the authors use stochastic differential equations (SDEs) to model the system, and, using a numerical solver, integrate in order to obtain a path, and thus a mapping. In contrast, we \emph{directly} learn the mappings and paths simultaneously. \Cref{tab:nlsb} compares the marginal distribution of the pushforwards from with fresh samples, and \cref{fig:synthetic-examples} contains the trajectories. In all scenarios, our mapping gives a more faithful estimate of the target distribution.

We want to stress that, even though we present paths, our Lagrangian OT maps are also learned and deterministic, whereas the NLSB method requires integrating and SDE to obtain the final displacement. Finally, NLSB presents two path variants: stochastic paths, and expected paths. The former is the trajectory of a bona fide SDE, and the latter is the average trajectory starting from a given point. On one hand, the expected path follows an ODE, and is easier to integrate; on the other, the trajectories are more degenerate, and are significantly less meaningful.

\begin{table}[t]
  \caption{Marginal $2$-Wasserstein errors \detail{(scaled by 100x)} of the push-forward measure on the synthetic settings from \citet{koshizuka2022neural}.}
  \label{tab:nlsb}
  \centering
  \resizebox{\linewidth}{!}{
  \begin{tabular}{r|llll}
    \toprule
    & box & slit & hill & well \\ \midrule
    NLOT (ours) & \cellhi $\bf 1.6 \pm 0.2$ & \cellhi $\bf 1.3 \pm 0.2$ & \cellhi $\bf 1.8 \pm 1.3$ & \cellhi $\bf 1.3 \pm 0.3$ \\
    NLSB (stochastic) & $2.4 \pm 0.6$ & \cellhi $\bf 1.3 \pm 0.4$ & \cellhi $\bf 2.0 \pm 0.1$ & $2.6 \pm 1.6$  \\
    NLSB (expected) & $6.0 \pm 0.5$ & $17.6 \pm 1.8$ & $4.0 \pm 0.9$ & $16.1 \pm 3.5$ \\
    \bottomrule
  \end{tabular}}
{\footnotesize\color{gray}
$^*$Results are from training three trials for every method.}
\end{table}

\subsection{Related work}
\paragraph{Lagrangian Schr{\"o}dinger bridges.}
\citet{koshizuka2022neural} and \citet{liu2022deep,liu2024generalized} are the closest for
this subproblem that we consider in \cref{sec:neural_lagot}.
The former studies the \emph{Stochastic Optimal Transport}
(SOT) problem, which amounts to optimal transport on path space, with
the path dictated by a stochastic differential equation
(SDE). The authors consider Lagrangian costs
and use neural SDEs to model the trajectories.
\citet{liu2022deep,liu2024generalized} investigate the generalized \emph{Schr{\"o}dinger Bridge Problem} (SBP),
which can be distilled to optimal transport with entropic
regularization \citep{leonard2012schrodinger}, also using neural
SDEs, and have a particular focus on mean-field games, which is not a focus of this work (see for example \citet{lin2020apac,ruthotto2020machine,
ding2022mean})
The SBP is also a special case of SOT;
see \citet[Figure 2]{koshizuka2022neural}. 

\paragraph{Estimation and applications of optimal transport maps under the squared-Euclidean cost.}
Apart from machine learning communities, the squared-Euclidean cost
has also garnered much interest in traditional domains. For example, statistical estimation of optimal transport maps for the squared-Euclidean cost started with \citet{hutter2021minimax}, followed swiftly by
\citet{deb2021rates,manole2021plugin,pooladian2021entropic}, to name a
few. Applications of optimal transport in machine learning often revolve around ``generative modeling", where the notion
of a learned transport map allows us to generate new samples from a
target measure from which we only have access to samples (\eg,
generating a new image). Examples of such works include
\citet{huang2021convex,finlay2020learning,finlay2020train,lipman2022flow,pooladian2023multisample,onken2021ot, rout2021generative, bousquet2017optimal,
balaji2020robust,seguy2017large,tong2023conditional}. The work of
\citet{schiebinger2019optimal} had a cascading effect in the machine
learning community, popularizing the ability to predict single-cell
genome expressions through optimal transport using limited
data. Examples include
\citet{bunne2021learning,bunne2022proximal,bunne2022supervised,lubeck2022neural}
and \citet{tong2020trajectorynet}.
\vspace{-5mm}
\paragraph{Estimation of optimal transport maps for other notions of cost.}
One can generalize the notion of a Brenier-map in closed form by
considering a specific family of cost functions. We call a convex cost
function \emph{translation invariant} if $c(x,y)
\defeq h(x-y)$ with $h : \Rd \to \R$ convex. To the best of our
knowledge, the pursuit of estimating such maps has been seldom, apart
from \citet{fan2021neural} and the recent works
\citet{cuturi2023monge, uscidda2023monge,klein2023learning}. Other applications include defining notions of optimal transport between datasets or
different spaces \citep{nekrashevich2023neural, alvarez2020geometric,
alvarez2018gromov}.

\section{Metric learning with NLOT}\label{sec:diff_metric_learning}

The following set of experiments is inspired from recent works such as \citet{tong2020trajectorynet,bunne2022proximal,zhang2022wassersplines,schiebinger2019optimal} that assume data is obtained as sparse pairs of sequences $\{\rho_i,\rho_{i+1}\}_{i=1}^{K-1}$ (such a setup arises in single-cell genomic profiling, for example). At the core of these methods is the crucial assumption that the space is Euclidean, which allows the authors to leverage various facts about optimal transport maps arising from convex costs. In contrast, \citet{scarvelis2022riemannian} considers the perspective that the data in fact arises from a non-Euclidean Riemannian manifold, where the underlying metric is given by some twisted inner product with respect to a positive definite matrix (recall \cref{ex:geodesic} in \cref{sec:background_lagot}).

\begin{algorithm*}[t]
  \caption{Metric learning with NLOT}
  \begin{algorithmic}
    \STATE \textbf{inputs:} measures $\{(\rho_i, \rho_{i+1})\}_{i=1}^{K-1}$,
    metric $A_\vartheta$,
    potentials $g_{\theta_i}$,
    $c$-transform predictors $y_{\phi_i}$, spline predictors $\varphi_{\eta_i}$,
    \WHILE {unconverged}
    \STATE update $\vartheta$ using $\nabla_\vartheta \ell_{\text{dual}}$
        (with the terms in \cref{eq:metric-derivative})
    \STATE update the OT approximation $\theta_i, \phi_i, \eta_i$ with
        an iteration of \cref{alg:lagrangian-ot}
    \ENDWHILE
    \STATE \textbf{return} optimal parameters $\vartheta$, $\theta_i$, $\phi_i$, $\eta_i$
  \end{algorithmic}
  \label{alg:metric-learning}
\end{algorithm*}

With this perspective in mind, we now consider optimal
transport problems where the ground-truth displacement is given
by geodesics induced by non-Euclidean
geometries, like \cref{eq:circle_path}. However, we crucially
\emph{do not} assume knowledge of the underlying positive-definite
matrix-valued function $A(\cdot)$ that induces the Riemannian geometry.
Our goal is to instead
\emph{learn} $A(\cdot)$ on the basis of sequential pairs of
probability measures $\{(\rho_{i},\rho_{i+1})\}_{i=1}^{K-1}$, as well as the final transportation mappings and paths.

Let $A_\vartheta$ be the neural network parameterization of a
positive-definite matrix, with the network weights given by
$\vartheta$.
The matrix-valued function $A_\vartheta$ then induces
the cost $c_\vartheta$
\vspace{-3mm}
\begin{align}
  c_\vartheta(x,y) \defeq \hspace{-1mm} \inf_{\gamma \in \cC(x,y)} \left\{ \int_0^1 \tfrac12 \|\dot{\gamma}_t\|^2_{A_\vartheta(\gamma_t)}\dd t \right\}.
\end{align}
Following \citet{scarvelis2022riemannian}, our goal is to learn a metric that results in a geometry
with a minimal OT cost, \ie,
$\min_{\vartheta} \ell_{\text{metric}}(\vartheta)$ where
\begin{align}
  \label{eq:problem_2}
  \ell_{\text{metric}}(\vartheta) \defeq \frac{1}{K} \sum_{i=1}^{K-1} \text{OT}_{c_\vartheta}(\rho_i,\rho_{i+1}).
\end{align}
We use the neural networks from \cref{sec:neural_lagot} to approximate
the OT maps, resulting in
\begin{align}
  \label{eq:problem_2_approx}
  \ell_{\text{metric}}(\vartheta)\approx \max_{\{\theta_i\}_{i=1}^{K-1}} \frac{1}{K} \sum_{i=1}^{K-1} \ell_{\text{dual}}(\theta_i; \rho_i, \rho_{i+1}, \vartheta).
\end{align}
Altogether, we aim to solve the following min-max optimization problem
\begin{align}
    \min_{\vartheta}\max_{\{\theta_i\}_{i=1}^{K-1}} \frac{1}{K} \sum_{i=1}^{K-1} \ell_{\text{dual}}(\theta_i; \rho_i, \rho_{i+1}, \vartheta)
\end{align}
with alternating descent-ascent.
For a fixed metric $A_\vartheta$, the inner maximization problem is
the same as \cref{sec:neural_lagot}, but with $K-1$ networks. The only
difference is the outer minimization step, which we compute
efficiently via sequential applications of the envelope theorem.
Noting that only the first term in \cref{eq:problem_2} depends on $A_\vartheta$,
the gradient of \cref{eq:problem_2_approx} is given by
\begin{align}
  \label{eq:metric-derivative}
  \begin{split}
    \nabla_\vartheta \ell_{\text{dual}}(\theta; \rho_i, \rho_{i+1}, \vartheta) &= \nabla_\vartheta \int g^{c_\vartheta} \dd \rho_i \\
    &\hspace{-10mm}=
    \int \nabla_\vartheta g^{c_\vartheta} \dd \rho_i \\
    &\hspace{-10mm}= \int \nabla_\vartheta c_\vartheta(x,\hat{y}(x)) \dd \rho(x)\\
    &\hspace{-10mm}= \int \nabla_\vartheta E_\vartheta(\varphi_{\eta_i}, x, \hat{y}(x))) \dd \rho_i\,.
  \end{split}
\end{align}
The full update of $A_\vartheta$ then takes the average gradient of these $K-1$ gradient
computations. Thus, the inner maximization step requires $K-1$
applications of \cref{alg:lagrangian-ot}, and the outer minimization
step freezes the inner parameters, leaving only an average update for
$A_\vartheta$. We stress that a primary difference in this setting
limited finite-sample access to the $K-1$ measures from which we are
to learn the ground-truth metric, and output paths and optimal
transport maps.
\Cref{alg:metric-learning} overviews the general algorithm.

\subsection{Experiments}\label{sec:experiments}

\begin{table*}[t]
  \centering
  \caption{Alignment scores $\ell_{\text{align}} \in [0,1]$
    for metric recovery in \cref{fig:learned-metrics}. \detail{(higher is better)}}\vspace{-2mm}
  \begin{tabular}{rlll}\toprule
    & Circle & Mass Splitting & X Paths \\\midrule
    Metric learning with NLOT (ours) & \cellhi $\bf 0.997 \pm 0.002$ & \cellhi $\bf 0.986 \pm 0.001$ & \cellhi $\bf 0.957 \pm 0.001$ \\
    \citet{scarvelis2022riemannian} & $0.995$ & $0.839$ & $0.916$ \\
  \end{tabular}
  \label{tab:scarvelis-alignment-scores}
\end{table*}

\begin{figure*}[t]
  \centering
  \hspace*{-5mm}
  \resizebox{.65\linewidth}{!}{
  \begin{tikzpicture}[every node/.style={align=left,anchor=west, inner sep=0}]
    \node[anchor=south west] at (0,0) {\includegraphics[width=0.82\textwidth]{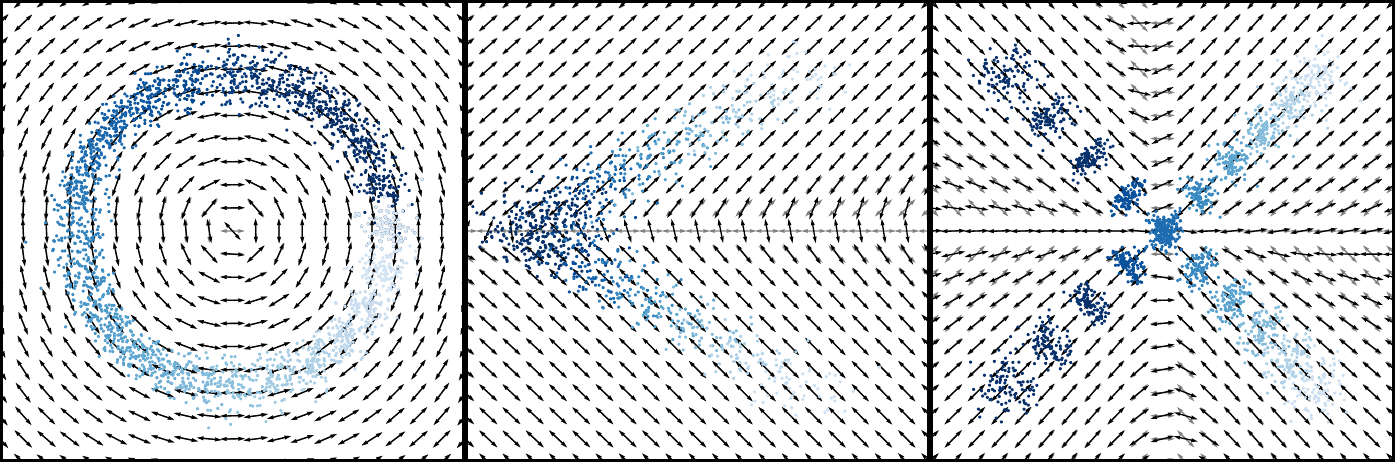}};
    \node[fill=white] at (1mm,1.75in) {Circle};
    \node[fill=white] at (1.9in,1.75in) {X Paths};
    \node[fill=white] at (3.7in,1.75in) {Mass splitting};
  \end{tikzpicture}}
  \hspace*{-5mm} \\
  \hspace*{-7mm} \detail{smallest} eigenvectors of $A$ {(\cblock{0}{0}{0} learned \cblock{127}{127}{127} ground-truth)}
  \cblock{38}{106}{177} data \detail{(lighter colors=later time)}
  \caption{We successfully recover the metrics on the
    settings from \citet{scarvelis2022riemannian}.}
  \label{fig:learned-metrics_data}
\end{figure*}
\begin{figure*}[t!]
  \centering
  \hspace*{-5mm}
  \resizebox{.65\linewidth}{!}{
  \begin{tikzpicture}[every node/.style={align=left,anchor=west, inner sep=0}]
    \node[anchor=south west] at (0,0) {\includegraphics[width=0.82\textwidth]{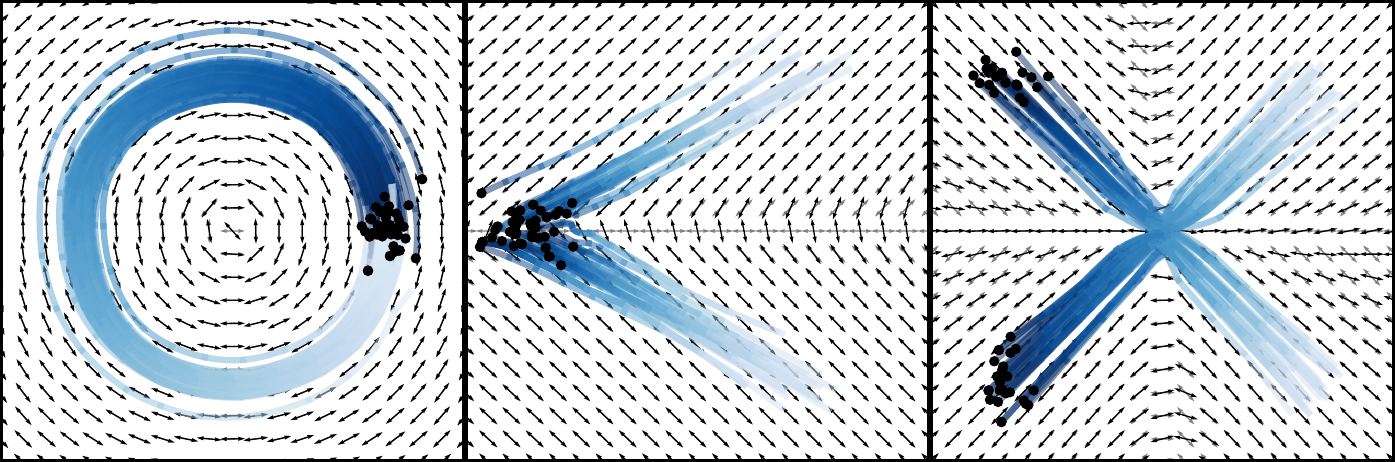}};
    \node[fill=white] at (1mm,1.75in) {Circle};
    \node[fill=white] at (1.9in,1.75in) {X Paths};
    \node[fill=white] at (3.7in,1.75in) {Mass splitting};
  \end{tikzpicture}}
  \hspace*{-5mm} \\
  \noindent \cblock{0}{0}{0} samples from initial measure
  \cblock{38}{106}{177} geodesics {\color{gray}(lighter colors=later time)}
  \caption{Our transport geodesics are able to reconstruct continuous versions of the
    original data that can predict the movement of individual particles
    given only samples from the first measure.
  }
  \label{fig:learned-metrics}
\end{figure*}

We consider three ground-truth Riemannian metrics $A(\cdot)$, which are given by the arrows in \cref{fig:learned-metrics} and \cref{fig:learned-metrics_data}.
To be precise, the grey arrows in the figures show the direction of the smallest eigenvector at that point \ie, the easiest direction to move in.
Note that \cref{fig:learned-metrics_data}(a) is the circle metric from \cref{eq:circle_path}, and the other two metrics are non-smooth metrics that cause splitting
\cref{fig:learned-metrics_data}(b) or reflections \cref{fig:learned-metrics_data}(c).
In each task, we are given samples from $K-1$ pairs of probability
measures which were generated according to these Riemannian metrics.
Our task is to learn the metric on the basis of the samples alone, and
ideally recover the transport path exactly.  The precise formulas for
$A(\cdot)$ and descriptions of the learning tasks are in
\cref{app:scarvelis_data}.

We parameterize the metric $A_\vartheta$ to predict a rotation
matrix $R_\vartheta(x)$ of a fixed matrix $B$, \ie,
\begin{equation}
  \label{eq:A-appendix}
  A_\vartheta(x) \defeq R_\vartheta(x) B R_\vartheta(x)^\top\,, \quad   B\defeq
  \begin{bmatrix}
    1 & 0 \\
    0 & 0.1 \\
  \end{bmatrix}\,,
\end{equation}
and the rotation matrix
\begin{equation}
  \label{eq:R}
  R_\vartheta(x)\defeq
  \begin{bmatrix}
    \cos\theta_\vartheta(x) & -\sin\theta_\vartheta(x) \\
    \sin\theta_\vartheta(x) & \cos\theta_\vartheta(x) \\
  \end{bmatrix}
\end{equation}
is obtained from predicting a rotation $\theta_\vartheta(x)$ from
the input $x$.
While this seemingly ad hoc approach appears limited to rotations, we find this to be far from the truth, as this method succeeds in learning three different geometries.
The eigenvalues of $A_\vartheta(x)$ are always the eigenvalues of $B$ ($1$ and $0.1$), and parameterizing the rotation forces $B$ to
be rotated so the data movement is along the smallest eigenvector
as in \cref{fig:learned-metrics_data}.

We quantify our ability to recover the ground-truth metric through the
alignment score from \citet{scarvelis2022riemannian}:
\begin{align}
  \small
  \hspace*{-2mm}
    \ell_{\text{align}}(A,\hat{A}) \defeq \frac{1}{d|\cD|}\sum_{x \in \cD}\sum_{i=1}^d|u_i(x)^\top \hat{u}_i(x)|\,, %
  \label{eq:alignment}
\end{align}
where $\cD$ is a finite discretization of the space, and $u_i(x)$
(resp. $\hat{u}_i(x)$) is the (unit) eigenvector with eigenvalue
$\lambda_i$ (resp. $\hat{\lambda}_i$) for the matrix $A(x)$
(resp. $\hat{A}(x)$). Our results are reported in
\cref{tab:scarvelis-alignment-scores}, where we perform the same
experiment over three randomized trials, and report the same metric
values from \citet{scarvelis2022riemannian}. Notably, we see a roughly
17\% improvement in the ``Mass Splitting" example, with near-perfect
recovery. Finally, in \cref{fig:learned-metrics}, we plot our fitted geodesics that are learned from the data. Unlike \citet{scarvelis2022riemannian}, our formulation allows us to output these geodesics, and does not require a separate training scheme; we elaborate on this point in the following subsection. 

\subsection{Related work}\label{sec:related work}
Although our setup is taken from \citet{scarvelis2022riemannian},
there are several differences between our work and theirs.
They deploy a specialized duality theory based on \cref{sec:background_kantdual},
where the Kantorovich potentials must be $1$-Lipschitz with respect to
the weighted Euclidean metric; this is enforced using an additional regularizer in the inner
maximization problem in \cref{eq:problem_2}.
Finally to ensure the metric does not collapse, they add another
regularizer on \cref{eq:problem_2}, for the outer minimization
problem.
They use \cref{eq:problem_2} \emph{only} to fit a metric
$\hat{A}_\vartheta$, and later use another optimization problem (based
on continuous normalizing flows) to fit their geodesics.
In contrast, our approach is self-contained, unregularized,
generalizable to other notions of cost, and we 
directly    obtain approximations of the transport
map with $y_\zeta$ and transport paths with $\varphi_\eta$.

\section{Conclusion}
In this work, we proposed an efficient framework for computing
geodesics under generalized least-action principles, or Lagrangians,
leading to large-scale computation of Lagrangian Optimal Transport
trajectories. Combining amortization and the use of splines, we
demonstrate the capacity of our method on a suite of problems, ranging
from learning non-Euclidean geometries from data, to computing optimal
transport maps under (known) non-Euclidean geometries and costs. There
are many remaining fundamental research avenues that arise as a result
of this work. Examples include a statistical analysis of these new
costs (\eg, \citet{hundrieser2023empirical,hutter2021minimax}),
extensions to the unbalanced optimal transport setting through the
Wasserstein-Fisher-Rao metric \citep{gallouet2017jko}, and extensions
to multi-marginal optimal transport \citep{pass2015multi}.

%
%
%
%
%
%
%
%
%

\bibliography{refs}

\newpage

\onecolumn
\title{Neural Optimal Transport with Lagrangian Costs\\(Supplementary Material)}
\maketitle
\appendix
\section{Details on the use of splines for
  geodesics and paths}\label{app:paths}

This section provides more background information and details
behind the amortized splines in \cref{sec:neural_lagot}
that address \cref{challenge:cost}.

\subsection{Cubic splines}
Cubic splines, \eg, as reviewed in \citet{mckinley1998cubic,wolberg1988cubic,bartels1995introduction,weisstein2008cubic,hastie2009elements,burden2015numerical},
are a widely-used method for fitting a parametric function to data.
We start with a review of general splines in one dimension (\cref{sec:1d-splines}),
then extend those to multiple dimensions (\cref{sec:nd-splines}),
then use those for representing the Lagrangian paths and geodesics
(\cref{sec:spline-geodesics}),
then amortize those (\cref{sec:amortized-spline-geodesics}).

\subsubsection{\ldots in one dimension}
\label{sec:1d-splines}
In one dimension, a cubic spline is defined by
\begin{equation}
  \gamma(x) =
  \begin{cases}
    \gamma_1(x) & \text{ if } x_1  \leq x < x_2 \\
    \gamma_2(x) & \text{ if } x_2  \leq x < x_3 \\
    \gamma_{n-1}(x) & \text{ if } x_{n-1}  \leq x < x_n \\
  \end{cases}
  \label{eq:cubic-spline-1d}
\end{equation}
where $x\in\R$, $x_i$ for $i\in\{1, \ldots, n\}$ are the \emph{knot points}
and
\begin{equation}
  \gamma_i(x)\defeq a_i+b_ix+c_ix^2+d_ix^3
  \label{eq:cubic-spline-component}
\end{equation}
are the cubic component functions with coefficients
$\bar\varphi_i\defeq[a_i, b_i, c_i, d_i]$.
We write the vector of all coefficients
as $\bar\varphi\defeq [\bar\varphi_1, \ldots, \bar\varphi_{n-1}]$.

\begin{challenge}[Parameterizing splines]
The coefficients $\bar\varphi$ are unknown and fit to data.
While they could be taken directly as the parameters for $\gamma$,
it would not result in a continuous function around the knot points.
\end{challenge}

The standard approach to resolve these discontinuities is to
constrain the component functions to be continuous and have
matching values and derivatives

\begin{equation}
  \begin{aligned}
  \gamma_i(x_{i+1})=\gamma_{i+1}(x_{x+1})\text{ for }i\in\{1,\ldots,n-1\} \\
  \gamma_i'(x_{i+1})=\gamma_{i+1}'(x_{x+1})\text{ for }i\in\{1,\ldots,n-1\} \\
  \gamma_i''(x_{i+1})=\gamma_{i+1}''(x_{x+1})\text{ for }i\in\{1,\ldots,n-1\}. \\
  \end{aligned}
  \label{eq:spline-conditions}
\end{equation}

These constraints, along with other conditions can be used to
provide a set of basis vectors $B\defeq[b_i]_{i=1}^m$ where $b_i\in|\bar\varphi|$ of spline
parameterizations $\bar\varphi$ that satisfy \cref{eq:spline-conditions},
\eg, as in \citet[Section 5.2.1]{hastie2009elements}.
In other words, any linear combination of the basis vectors $b_i$
will result in a valid parameterization.
We can thus reparameterize the spline with $\varphi\in\R^m$ to be
based on linear combinations of the basis, providing
\begin{equation}
  \bar\varphi = B\varphi = \sum b_i\varphi _i
  \label{eq:spline-basis-param}
\end{equation}
The advantage of this reparameterization is that $\varphi$
is a parameterization of splines in the unconstrained reals
and can therefore be treated as a standard learnable parameter
for our geodesic computations.

\subsubsection{\ldots in multiple dimensions}
\label{sec:nd-splines}
The standard extension of splines to functions of multiple dimensions,
\eg, for graphics \citep{bartels1995introduction},
is to parameterize a one-dimensional spline \cref{eq:cubic-spline-1d}
on each coordinate.
We will notate these as $\gamma_\varphi: \R\rightarrow\R^d$,
$\gamma_\varphi(x)\defeq [\gamma_{\varphi_1}(x), \ldots, \gamma_{\varphi_d}(x)]$
where $\varphi_i$ is the parameterization of the basis coefficients
for each one-dimensional spline.

\subsection{Cubic splines for geodesics and Lagrangian paths}
\label{sec:spline-geodesics}
We follow \citet{beik2021learning,stochman} and represent geodesics
and Lagrangian paths between two points
$x,y$ by a multi-dimensional spline $\gamma_\varphi(t)$ parameterized
by $\varphi$ where $t\in[0,1]$ is the time.
The basis for the splines enforce the smoothness properties in
\cref{eq:spline-conditions} as well as the boundary conditions
$\gamma_\varphi(0)=x$ and $\gamma_\varphi(1)=y$.

\subsection{Amortized cubic splines for geodesics}
\label{sec:amortized-spline-geodesics}
Instead of computing the spline parameters $\varphi$ individually for
every geodesic, we propose to \emph{amortize} them across the
geodesics needed for the OT maps.
This results in parameterizing an amortization model $\varphi_\eta(x,y)$
that predicts the spline parameters for a geodesic between $x$
and $y$ that we learn with objective-based amortization
in \cref{eq:spline-amor}.

\section{Synthetic data for Lagrangians with potentials}\label{app:potential_data}
We consider five potential functions $U(x)$. The following four potential functions are from \citet{koshizuka2022neural}:
\begin{align}
    & U_{\text{box}}(x) \defeq -M_1 \cdot \bm{1}_{[-0.5,0.5]^2}(x)\,,\\
    & U_{\text{slit}}(x) \defeq -M_2 \cdot ( \bm{1}_{([-0.1,0.1],(-\infty,-0.25] )}(x_1,x_2) + \bm{1}_{([-0.1,0.1],[0.25,\infty) )}(x_1,x_2)  )\,, \\
    & U_{\text{hill}}(x) \defeq -M_3\|x\|^2\,,\\
    & U_{\text{well}}(x) \defeq -M_4\exp(-\|x\|^2)\,,
\end{align}
where $M_1,M_2,M_3$ and $M_4$ are constants.

The Gaussian-mixture example is taken from \citet{liu2022deep}, which amounts to the following potential function
\begin{align}
    U_{\text{GMM}}(x) \defeq -M_5\sum_{i=1}^3 \bm{1}_{B_i}(x)\,,
\end{align}
where $B_i \defeq \{x : \|x - m_i\|\leq 1.5\}$ with $m_i \in\{(6,6),(6,-6),(-6,-6)\}$.

We approximate the hard constraints using sigmoid functions. We make the choices $M_1 = 0.01$, $M_2 = 1$, $M_3 = 0.05$, $M_4 = 0.01$, $M_5 = 0.1$ --- we are unable to use the same choice of $M$ for all potentials as a result of numerical instabilities that arise in the geodesic computation.

\section{Data from \cite{scarvelis2022riemannian}}
\label{app:scarvelis_data}
We briefly outline the three datasets used in \cref{sec:diff_metric_learning}, all of which were taken directly from \citet{scarvelis2022riemannian}, following their open source repository \url{https://github.com/cscarv/riemannian-metric-learning-ot}; here we simply explain the data generating processes.

The three datasets have a similar flavor: Let $\gamma$ be a time-varying curve, and suppose we have access to the matrix function $A(\cdot)$ which generates the known geometry. This allows the authors to generate a velocity field between two fixed points $x$ and $y$ (respectively, initial and final position of $\gamma$) using the following optimization problem
\begin{align}\label{eq: scarvelis_opt_app}
\min_\theta \int_0^1 \| v_{(t,\theta)}(\gamma_t) \|_{A(\gamma_t)}^2 \dd t + \| \gamma(1) - y\|   \,,
\end{align}
where $v_{(t,\theta)}(\cdot)$ is a time-varying neural network (parametrized by $\theta$) that is the solution to a neural ODE, where they also enforce the initial condition $\gamma_0 = x$. The integral in time is replaced with a sum over  indices $0 = t_1 < t_2 < \ldots < t_m = 1$. For a given collection of samples from measures $\{\rho_i\}_{i=1}^{K-1}$, the authors randomly pair up the data and solve \cref{eq: scarvelis_opt_app} across batches using the \citet{torchdiffeq} package (specifically using \texttt{odeint}). \Cref{eq: scarvelis_opt_app} is solved using AdamW with a learning rate of $10^{-3}$ and weight-decay factor $10^{-3}$, with 100 epochs of training per pair of samples. The learned solution $v_{(t,\hat{\theta})}$ is able to generate data at various time-points. With this setup in mind, we can turn to precise details for the three datasets.

\paragraph{Circular trajectory}
The circular path is enforced using the matrix
\begin{align}\label{eq:circle_path_app}
    A(x) = \begin{pmatrix}
        \frac{x_1^2}{\|x\|^2} & 1 - \frac{x_1x_2}{\|x\|^2} \\
        1 - \frac{x_1x_2}{\|x\|^2} & \frac{x_2^2}{\|x\|^2}
    \end{pmatrix} \,.
\end{align}
The goal is to generate Gaussian data that flows according to $A$. To this end, the authors fix four possible means (in order)  $\mu \in \{(1,0), (0,1), (-1,0), (0,-1)\}$, and fix $\sigma \defeq 0.1$, which define $\rho_i \defeq N(\mu_i, \sigma^2)$. 100 samples are drawn from each $\rho_i$, which constitutes the finite-samples that are used in the objective function \cref{eq: scarvelis_opt_app}. Once the velocity field is learned, there are 24 equispaced time-points from which they draw samples, resulting in 24 Gaussian distributions that flow according to $A$.
\looseness=-1

\paragraph{Mass-splitting trajectory}
In this example, $A(x) = I - w(x)w^\top(x)$, with
\begin{align}
    w(x) =
    \begin{cases}
        \left(\tfrac{1}{\sqrt{2}},\tfrac{1}{\sqrt{2}}\right) & x_2 \geq 0\,,\\
        \left(\tfrac{1}{\sqrt{2}},\tfrac{-1}{\sqrt{2}}\right) & x_2 < 0\,.
    \end{cases}
\end{align}
In this case, there are three Gaussians, with means $\mu_i \in \{(0,0), (10,10), (10,-10)\}$ and unit variance. Again, 100 samples are drawn from each, which are randomly paired and allow the authors to numerically solve \cref{eq: scarvelis_opt_app}. Once they have a learned vector field, they generate the data at 10 equispaced time-points.

\paragraph{X-path trajectory}
In this third case example, $A(x) = I - w(x)w^\top(x)$, with
\begin{align}
    w(x) = \alpha(x)w_1(x) + \beta(x)w_2(x)
\end{align}
where $w_1(x) = (\tfrac{1}{\sqrt{2}},\tfrac{1}{\sqrt{2}})$ and
$w_2(x) = (\tfrac{1}{\sqrt{2}},\tfrac{-1}{\sqrt{2}})$,
and
$\alpha(x) = 1.25\tanh(\text{ReLU}(x_1x_2)$ and $\beta(x)= 
-1.25\tanh(\text{ReLU}(-x_1x_2)$. Here, there are two sets of two
trajectories, corresponding to Gaussian data with means $\mu_i^{(1)}
\in \{(-1,-1),(1,1)\}$ and $\mu_i^{(2)} \in \{(-1,1),(1,-1)\}$, all
with standard deviation $\sigma= 0.1$. As before, 100 samples are
generated, and \cref{eq: scarvelis_opt_app} is solved (twice)
numerically; 10 time-points per velocity field are used to generate
the total data.

\section{Hyper-parameters}\label{sec:training_specs}

\begin{table}[H]
  \caption{Hyper-parameters for computing the OT maps
    in \cref{fig:synthetic-examples_intro,fig:synthetic-examples}
    with \cref{alg:lagrangian-ot}.}
  \label{tab:hypers}
  \centering
  \begin{tabular}{r|l}
    \toprule
    Hyper-Parameter & Value \\ \midrule
    Number of spline knots & $30$ \\
    $g_\theta$ MLP layer sizes & \verb![64, 64, 64, 64]! \\
    $y_\zeta$ MLP layer sizes & \verb![64, 64, 64, 64]! \\
    $\gamma_\varphi$ MLP layer sizes & \verb![1024, 1024]! \\
    MLP activations & Leaky ReLU \\
    $g_\theta$ learning rate schedule &
      Cosine \detail{(starting at $10^{-4}$ and annealing to $10^{-2}$)} \\
    $y_\zeta$ learning rate schedule &
      Cosine \detail{(starting at $10^{-4}$ and annealing to $10^{-2}$)} \\
    $\gamma_\varphi$ learning rate & $10^{-4}$ \detail{(no schedule)} \\
    Batch size & $1024$ \\
    $c$-transform solver &
      LBFGS \detail{($20$ iterations, backtracking Armijo line search)} \\
    \bottomrule
  \end{tabular}
\end{table}

\begin{table}[H]
  \caption{Hyper-parameters for computing
  \cref{fig:learned-metrics_data,fig:learned-metrics,tab:scarvelis-alignment-scores}
  with \cref{alg:lagrangian-ot} and \cref{eq:problem_2}.}

  \label{tab:hypers}
  \centering
  \begin{tabular}{r|l}
    \toprule
    Hyper-Parameter & Value \\ \midrule
    Number of spline knots & $30$ \\
    $g_\theta$ MLP layer sizes & \verb![64, 64, 64, 64]! \\
    $y_\zeta$ MLP layer sizes & \verb![64, 64, 64, 64]! \\
    $\gamma_\varphi$ MLP layer sizes & \verb![1024, 1024]! \\
    MLP activations & Leaky ReLU \\
    $g_\theta$, $y_\zeta$, $\gamma_\varphi$ learning rates &
       $10^{-4}$ \detail{no schedule} \\
    Batch size & $1024$ \\
    $c$-transform solver &
      LBFGS \detail{($20$ iterations, backtracking Armijo line search)} \\ \midrule
    $A_\vartheta$ learning rate & $5\cdot 10^{-3}$ \\
    Update frequency & $1$ update of $A_\vartheta$ for
            every $10$ updates of $g_\theta$, $y_\zeta$, and $\gamma_\varphi$ \\
    \bottomrule
  \end{tabular}
\end{table}

\end{document}

%% file: paper_preamble.tex
\usepackage{microtype}
\usepackage{graphicx}
\usepackage{subfigure}
\usepackage{booktabs}
\usepackage{xspace}
\usepackage[table,svgnames]{xcolor}

\usepackage{amsmath,amssymb,amsthm, mathtools}
\theoremstyle{plain}

\theoremstyle{definition}

\theoremstyle{remark}
\newtheorem{remark}{Remark}
\theoremstyle{example}
\newtheorem{example}{Example}
\newtheorem{challenge}{Challenge}

\usepackage{wrapfig}
\usepackage{enumitem}
\usepackage{bm}

\definecolor{linkcolor}{RGB}{74, 102, 146}
\usepackage[
colorlinks=true,allcolors=linkcolor,pageanchor=true,
plainpages=false,pdfpagelabels,bookmarks,breaklinks,bookmarksnumbered,
backref=page,
]{hyperref}

\usepackage{mathtools,amsmath,amsthm, amssymb}
\usepackage{xifthen}
\usepackage{dsfont}

\newcommand{\cC}{\mathcal{C}}
\newcommand{\cD}{\mathcal{D}}

\newcommand{\cL}{\mathcal{L}}
\newcommand{\cM}{\mathcal{M}}

\newcommand{\cP}{\mathcal{P}}

\newcommand{\cX}{\mathcal{X}}
\newcommand{\cY}{\mathcal{Y}}

\newcommand*{\kl}[3][]{%
\ifthenelse{\isempty{#1}}{\operatorname{KL}(#2\,\|\,#3)}%
{\operatorname{KL}(#2\,\|\,#3\mid#1)}%
}

\newcommand*{\triplenorm}[1]{{\left\vert\kern-0.25ex\left\vert\kern-0.25ex\left\vert #1
    \right\vert\kern-0.25ex\right\vert\kern-0.25ex\right\vert}}

\newcommand*{\defeq}{\coloneqq}
\newcommand*{\rd}{\mathrm{d}}
\newcommand*{\dd}{\, \rd}
\DeclareMathOperator*{\argmin}{argmin}

\newcommand{\R}{\mathbb{R}}
\newcommand{\Rd}{\mathbb{R}^d}

\newcommand{\ie}{\emph{i.e.}\xspace}
\newcommand{\eg}{\emph{e.g.}\xspace}
\newcommand{\cf}{\emph{cf.}\xspace}

\usepackage{todonotes}

\newcommand{\cellhi}{\cellcolor{RoyalBlue!15}}

\definecolor{detailcolor}{RGB}{120, 120, 120}
\newcommand{\detail}[1]{{\color{detailcolor} #1}}

\usepackage[nameinlink]{cleveref}
\Crefname{algorithm}{Alg.}{Algs.}
\Crefname{challenge}{Challenge}{Challenges.}

\newcommand{\cblock}[3]{
  \hspace{-1.5mm}
  \begin{tikzpicture}[node/.style={square, minimum size=10mm, thick, line width=0pt}]
    \node[fill={rgb,255:red,#1;green,#2;blue,#3}] () [] {};
  \end{tikzpicture}%
}